\documentclass[journal,twoside,web]{ieeecolor}
\usepackage{generic}
\usepackage{cite}
\usepackage{amsmath,amssymb,amsfonts}
\usepackage{algorithmic}
\usepackage{graphicx}
\usepackage{textcomp}

\usepackage{booktabs}  
\usepackage{algorithm}
\usepackage{cite} 

\usepackage{hyperref}
\hypersetup{
		hypertex=true, 
		colorlinks=true, 
		linkcolor=blue, 
		anchorcolor=blue, 
		citecolor=blue
	}

\def\BibTeX{{\rm B\kern-.05em{\sc i\kern-.025em b}\kern-.08em
    T\kern-.1667em\lower.7ex\hbox{E}\kern-.125emX}}
\begin{document}
\title{Towards Real-World Applications of Personalized Anesthesia Using Policy Constraint Q Learning for Propofol Infusion Control}
\author{Xiuding Cai, Jiao Chen, Yaoyao Zhu, Beimin Wang, Yu Yao
	\thanks{This work was supported in part by the National Natural Science Foundation of China under Grant 82073338, and in part by the Sichuan Provincial Science and Technology Department under Grant 2022YFS0384 and 2022YFQ0108. (Corresponding authors: Jiao Chen.)}
	\thanks{Xiuding Cai, Yaoyao Zhu, Beiming Wang and Yu Yao are with Chengdu Institute of Computer Application, Chinese Academy of Sciences, Chengdu, China, and with the School of Computer Science and Technology, University of Chinese Academy of Sciences, Beijing, China (e-mail: \{caixiuding20, zhuyaoyao19, wangbeimin21\}@mails.ucas.ac.cn, casitmed2022@163.com). }
	\thanks{Jiao Chen was with Department of Anesthesiology, West China Hospital, Sichuan University \& The Research Units of West China (2018RU012), and with Chinese Academy of Medical Sciences, China (e-mail: chenjiao@wchscu.cn).}}

\maketitle

\begin{abstract}
Automated anesthesia promises to enable more precise and personalized anesthetic administration and free anesthesiologists from repetitive tasks, allowing them to focus on the most critical aspects of a patient's surgical care. Current research has typically focused on creating simulated environments from which agents can learn. These approaches have demonstrated good experimental results, but are still far from clinical application. In this paper, Policy Constraint Q-Learning (PCQL), a data-driven reinforcement learning algorithm for solving the problem of learning strategies on real world anesthesia data, is proposed. Conservative Q-Learning was first introduced to alleviate the problem of Q function overestimation in an offline context. A policy constraint term is added to agent training to keep the policy distribution of the agent and the anesthesiologist consistent to ensure safer decisions made by the agent in anesthesia scenarios. The effectiveness of PCQL was validated by extensive experiments on a real clinical anesthesia dataset we collected. Experimental results show that PCQL is predicted to achieve higher gains than the baseline approach while maintaining good agreement with the reference dose given by the anesthesiologist, using less total dose, and being more responsive to the patient's vital signs. In addition, the confidence intervals of the agent were investigated, which were able to cover most of the clinical decisions of the anesthesiologist. Finally, an interpretable method, SHAP, was used to analyze the contributing components of the model predictions to increase the transparency of the model.
\end{abstract}


\section{Introduction}
\label{sec:introduction}
\IEEEPARstart{A}{nesthesia} is a critical component of the operating room, with millions of patients requiring general anesthesia during surgery each year~\cite{weiser2008estimation}. Anesthesiologists face tremendous work pressure every day. They must monitor the patient throughout the procedure while maintaining several aspects of the patient simultaneously, including anesthesia status, physiological stability, pain management, and oxygen delivery. Routinely, an anesthesiologist is responsible for the life support of more than one patient at a time. However, the increasing number of operations in recent years has posed a significant challenge to the limited number of anesthesiologists~\cite{kempthorne2017wfsa}. This serious challenge not only puts immense pressure on anesthesiologists, but also increases the risk of medical malpractice and burnout~\cite{afonso2021burnout}. Given the critical role of anesthesia, the repetitive nature of anesthesiologists' tasks, and the shortage of positions, automated anesthesia infusion has emerged as an essential research direction in healthcare, offering a promising solution to alleviate the	 challenge. Automatic anesthetic administration has been shown to allow for more accurate and responsive anesthetic drug control with a reduced total dose than complete manual control by the anesthesiologist~\cite{puri2016multicenter, brogi2017clinical, zaouter2020autonomous, ghita2020closed}. This approach also helps reduce the patient's postoperative recovery time, as high doses are currently known to substantially increase the likelihood of side effects, such as propofol infusion syndrome~\cite{pasin2017closed}.

Reinforcement learning (RL) is a subfield of machine learning that aims to solve sequential decision problems. RL interacts with the environment by creating an agent that obtains observational states and rewards from the environment to adjust the policy to maximize cumulative rewards and further achieve optimal control. In recent years, RL has been successfully applied in healthcare, including breast cancer screening~\cite{yala2022optimizing}, sepsis treatment~\cite{raghu2017continuous}, glioblastoma treatment~\cite{zade2020reinforcement}, diabetes glucose control~\cite{zhu2020basal, noaro2023personalized}, etc. Anesthesia infusion can be viewed as a decision-making process over time, in which the anesthesiologist selects the optimal combination of drug dosages based on the patient's clinical information, as well as current physical status (\emph{e.g.}, blood pressure, heart rate, depth of anesthesia, etc.) to maintain the patient's vital signs within the target interval. Therefore, the anesthesia infusion problem is naturally amenable to modeling using RL, and there is a promise for using RL for more efficient and effective anesthesia administration. 

\begin{figure*}[!t]
	\centerline{\includegraphics[width=0.75\linewidth]{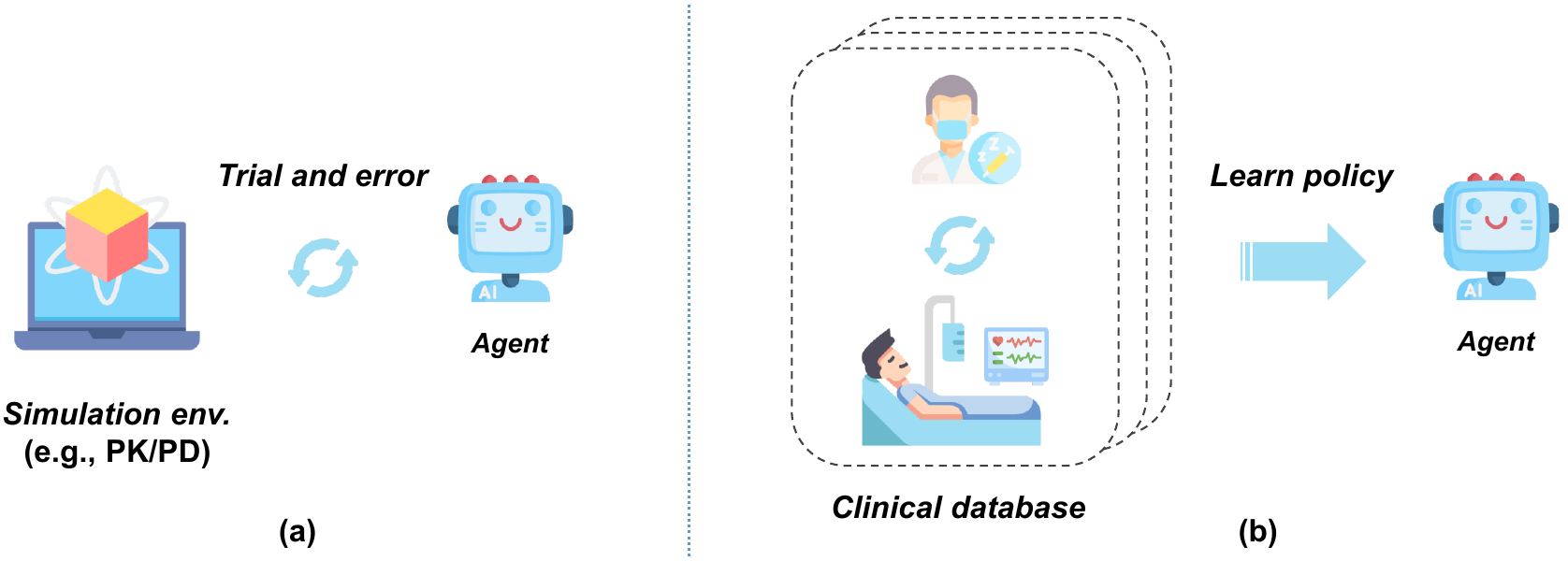}}
	\caption{Comparison of traditional RL (the left) and offline RL (the right) for automatic anesthesia policies learning.}
	\label{fig:rl_vs_orl}
\end{figure*}

However, traditional RL learns optimal policies by interacting with the environment through trial and error (see Fig.~\ref{fig:rl_vs_orl}). However, anesthesia is a real clinical procedure, and any dose misuse can harm the patient. For instance, overdosing propofol can damage the patient's brain, and underdosing can cause unnecessary intraoperative awareness. It is inhumane and high-risk to train RL agents directly with the human body. To this end, Moore et al.~\cite{moore2011reinforcement} proposed pharmacokinetic/pharmacodynamic (PK/PD) models, which simulate the distribution and transfer of drugs in the human body through differential equations. However, such models traditionally involve many hyperparameters, which are usually derived from population statistical data, which means that such models ignore the specificity across patients and therefore may be highly inaccurate when modeling an individual's environment. By means of personalized multi-task learning, Bird et al.~\cite{bird2019multi} improved the PK/PD model. While such methods have proven successful experimentally, simulation-based learning remains distant from clinical applications. Recently, Shin et al.~\cite{shin2021joint} proposed to learn a function that simulates the human environment based on real-world clinical anesthesia surgery data collected. The agent is then trained in this environment. All these algorithms mentioned above rely on environment modeling, and the reliability of trained agents depends on the simulation credibility.

Recently, offline reinforcement learning (ORL) has achieved impressive success and has received increasingly widespread attention. ORL is an advanced RL paradigm in which agents can learn suboptimal or even optimal policies from a collected offline dataset, while not requiring additional environmental exploration. Given that no interaction with the environment required, this new RL paradigm has attracted considerable interest in healthcare~\cite{kondruppersonalization, emerson2022offline, shiranthika2022supervised, wang2022learning}. However, current ORL methods suffer from the problem of distributional shift~\cite{fujimoto2019off}. For example, it is easy to learn overestimated Q functions, assigning high Q values to unseen operations, thus causing the model to select uncommon actions in the dataset, leading to undesired results. Moreover, the anesthesia task places higher safety requirements on the learned RL algorithms than common game tasks such as Atari~\cite{bellemare2013arcade}. An abnormal action by the agent may endanger the patient's life. To this end, we propose Policy Constraint Q-Learning (PCQL), a data-driven RL algorithm for solving the problem of learning anesthesia infusion policies on real clinical datasets. To the best of our knowledge, we are the first to apply ORL in the context of real clinical anesthesia. We first learn a superior anesthesia infusion policy from the collected anesthesia dataset based on Conservative Q-Learning (CQL;~\cite{kumar2020conservative}), a class-leading ORL algorithm. To ensure safer decisions by the agent, we add a policy constraint regularization term during the learning of the agent policy. The regularizer is learnable and obtained by supervised training on a collected offline dataset, thus implicitly modeling the distribution of state-action pairs.

We conducted extensive experimental validation using a large clinical anesthesia dataset that we collected. Initially, we evaluated PCQL using the Off-Policy Evaluation method commonly employed in ORL. Our results suggested that PCQL outperformed all baselines, including human anesthesiologists' policies and other RL methods. Subsequently, we evaluated different RL algorithms trained on retrospective data considering the clinical dose by anesthesiologists as the reference dose. Owing to the policy constraint regularization term, PCQL agrees better with the anesthesiologist's policy than other RL baselines and has the lowest error in both MAPE and RMSE metrics. We further analyzed PCQL's dose usage and observed that it required lower total drug doses to maintain patients' vital signs within the target interval compared to the anesthesiologist's clinical strategy. Moreover, PCQL's recommended dose exhibited a stronger correlation with patient vital sign information and a faster adjustment frequency than the anesthesiologist's policy, indicating the potential for accurate and personalized automated anesthesia. We also analyzed PCQL's confidence intervals and found that they encompassed most anesthesiologists' policies, thereby bolstering the model's credibility and reliability. Finally, we utilized SHAP, an interpretable method, to analyze the model's decision results and enhance the transparency of the automated anesthesia control system, which is pertinent for anesthesiologists' utilization of the system.

\section{Related Works}
\label{sec:related_works}
\subsection{Automated anesthesia infusion}
Automated anesthesia technology was first pioneered in the 1980s and has since advanced even further, as well as becoming more widely used~\cite{Wingert2021MachineLD}. This technology frees anesthesiologists from repetitive drug control tasks, allowing them to focus on the most critical aspects of each case, resulting in high-quality care for each patient. Moreover, automated anesthesia has shown great promise in improving drug infusion regimens and allowing for better precision anesthesia. Pasin et al.~\cite{pasin2017closed} have shown that Bispectral-guided total intravenous anesthesia can reduce the need for propofol during induction, better maintain the target depth of anesthesia, and reduce recovery time compared to manual control. In a multicenter study, Puri et al.~\cite{puri2016multicenter} showed that automated anesthesia can consistently achieve better performance than manual control. In a retrospective study, Brogi et al.~\cite{brogi2017clinical} found that automated anesthesia can effectively reduce overshooting or undershooting of target physiological indicators and can effectively increase the duration of maintenance at the target interval. The current development of automated anesthesia infusion technology is usually closely related to the evolution of artificial intelligence. Moore et al.~\cite{moore2011reinforcement} used a proportional integral derivative (PID) controller for controlling propofol infusion rate, which in turn regulates Bispectral (BIS). Some other control algorithms for automatic anesthesia have also been investigated, such as model-predictive controllers~\cite{sawaguchi2008model}, and rule-based controllers~\cite{liu2006titration}. However, such traditional control methods are usually limited by linear assumptions. For this reason, Moore et al.~\cite{moore2011reinforcement} first proposed the use of discrete action of RL for anesthesia control and achieved better results than PID controllers. Subsequently, Lowery et al.~\cite{lowery2013towards} extended the work of~\cite{moore2011reinforcement} to continuous space. Schamberg et al.~\cite{schamberg2022continuous} used a more advanced actor-critic algorithm to train an agent for anesthesia control. Yun et al.~\cite{yun2022hierarchical} used a hierarchical RL algorithm to learn a high-level and a low-level policy. The high-level policy generates a target BIS trajectory, and the low-level policy uses this information to learn more stable infusion control.

\subsection{Environmental simulation modeling of anesthesia}
Environment modeling is fundamental to RL because the agent interacts with the environment through a "trial-and-error" paradigm, from which it learns the target policy. Existing automated anesthesia algorithms typically rely on the development of pharmacokinetic (PK) and pharmacodynamic (PD) models that simulate the response of a patient's BIS level to a specific drug dose. The PK model describes how the drug flows through the various compartments of the body (\emph{e.g.}, brain, slow compartments, etc.) based on a system of equations for a particular drug. The PD model then maps the specific drug concentration at the effect site to the effect level, \emph{i.e.}, BIS. However, simulation-based human environments usually involve a wide range of hyperparameters, which are usually derived from the statistical values of the population. This means that such models ignore the specificity between different patients and may therefore suffer unexpected inaccuracies when simulating the human environment. Bird et al.~\cite{bird2019multi} uses multi-task learning techniques to personalize the PK/PD model to an individual level, while retaining statistical power, and the results show improved prediction accuracy. Despite the promising experimental results achieved by such methods, the simulation data are still not convincing for applications in clinical anesthesia. Several studies have attempted to learn on real-world collections of clinical anesthesia procedures, and the goal of such methods is to learn an environmental function that simulates the body's response after receiving a certain dose of medication. Shin et al.~\cite{shin2021joint} learned a transition function that simulates the human environment through supervised learning, and then used proximal policy optimization combined with behavioral cloning algorithms to learn automatic anesthesia policies based on this environment, and achieved promising experimental results. Unlike~\cite{shin2021joint}, our goal is to use real clinical data and train a scoring function that evaluates the plausibility of state-action pairs and thus guarantees safer decisions by the agent.

\subsection{Offline Reinforcement Learning}
Recently, an advanced RL paradigm, offline reinforcement learning (ORL), also known as batch reinforcement learning, has been proposed. Compared to traditional online reinforcement learning, ORL allows agents to learn superior policies from collected datasets without having to perform additional exploration in the environment. Given the property of "offline", and thus avoiding costly and dangerous, unethical exploratory actions when interacting with the environment, ORL has also gained widespread interest in the medical field. Kondrup et al.~\cite{kondruppersonalization} used deep conservative reinforcement learning to determine the best ventilator settings for ICU patients. Emerson et al.~\cite{emerson2022offline} proposed using ORL to learn a safer blood glucose control strategy for people with Type 1 diabetes. Shiranthika et al.~\cite{shiranthika2022supervised} developed the supervised optimal chemotherapy regimen, which can provide cancer patients with an optimal chemotherapy-dosing schedule, thus assisting oncologists in clinical decision-making. Wang et al.~\cite{wang2022learning} used ORL to learn the optimal treatment strategy for sepsis patients in ICU. These are good illustrations of the potential of ORL applications in the medical field. However, the current ORL is susceptible to the problem of distributional shift. For example, it is easy to learn overestimated Q functions that assign high Q values to unseen operations, thus causing the model to select uncommon operations in the dataset, putting patients at risk. Researchers have made successive efforts to alleviate the overestimation problem, such as Batch Constrained Q-Learning~\cite{fujimoto2019off}, Advantage Weighted Actor-Critic~\cite{nair2020awac}, Conservative Q-Learning~\cite{kumar2020conservative}, etc.

\subsection{Conservative Q-Learning}
Conservative Q-Learning (CQL;~\cite{kumar2020conservative}) is an ORL algorithm based on a Soft Actor-Critic improvement, which solves the problem of overestimating Q values in ORL by learning a conservative estimate of the Q function.

In a standard Q function, the loss function can be written as:
$$
\begin{aligned}
	L_{DQN}&=\mathbb{E}_{s, a \sim D}\left[\left(Q(s, a)-B^{\pi_{k}} Q^{k}(s, a)\right)^{2}\right]\\
	&=\mathbb{E}_{s, a \sim D}\left[\left(Q(s, a)-\left(r(s,a)+\gamma \max_{a'}Q(s', a')\right)\right)^{2}\right],
\end{aligned}
$$
where $B^{\pi_k}$ is the Bellman operator~\cite{sutton2018reinforcement} on the currently learned policy $\pi_k$ at iteration $k$. The true Q-value estimate of the state-action pair is approximated by minimizing $Q(s, a)-(r(s,a)+\gamma \max_{a'}Q(s', a')$. However, due to the inability to explore the environment, naive Q-learning tends to get overly optimistic Q values on offline data, which leads to the problem of overestimation. To this end, CQL adds a regularization term:
\begin{equation}
	\begin{aligned}
		L_{CQL}
		&=L_{DQN}\\
		&+\alpha \mathbb{E}_{s \sim D}\left[\log \sum_a \exp (Q(s, a))-\mathbb{E}_{a \sim D}[Q(s, a)]\right],
	\end{aligned}\label{cql}
\end{equation}
where $\alpha$ is the factor that relaxes the importance of the conservative term in the overall loss. The log-sum-exp term penalizes the action with the largest Q-value. The second term, on the other hand, guarantees to maximize Q-values for state-action pairs in the dataset. Thus, this conservative term allows high Q-values to be assigned only to actions within the distribution.

\section{Methodology}
\label{sec:methodology}

We first formalize the procedure of automatic anesthetic administration, including the state space, action space, and reward design. Subsequently, we introduce a new constraint term that enables to constrain agent-predicted actions in the action distribution of the dataset. Finally, we describe the proposed overall framework and the specific training process.

\subsection{The Problem Setting}

In this study, the automatic administration of anesthesia is modeled as a Markov decision process (MDP) at finite time steps. Typically, the MDP is defined as a 5-tuple $<S,A,R,P,\gamma>$. At time step $t$, the agent, in the current state $s_t\in S$, takes an action $a_t\in A$ and moves to the next state $s_{t+1}\in S$ according to the transition probability $P(s'|s,a)$. The agent expects to maximize the accumulated reward as it interacts with the environment.

In the problem setting, the agent is the controller that controls the delivery rate of the drug delivery device, while the environment is the patient in the perioperative phase of anesthesia. The agent is expected to predict the optimal dosage required by the patient at the current moment, based on the state observed from the environment, such as the patient's current vital signs information, and the history of drug administration. A suitable drug delivery controller should be able to maintain the patient's vital signs consistently during the surgery with minimal drug administration costs. Below are detailed definitions of the observed states, environment, reward functions, and agents.

\subsubsection{State Space}
The state space defines the information that the agent can observe at each moment, including the patient's clinical information, real-time vital signs, fluids, and other important information. The state space contains a total of 19 variables.
\begin{itemize}
	\item Clinical information: age, gender, height, weight, BMI, ASA grade.
	\item Vital Signs: systolic arterial pressure ($\text{AP}_\text{sys}$), diastolic arterial pressure ($\text{AP}_\text{dia}$), mean arterial pressure (MAP), MAP/$\text{AP}_\text{sys}$/$\text{AP}_\text{dia}$ for the previous two moments.
	\item Analgesics: Sufentanil.
	\item The others: MAP target, MAP change, MAP target error.
\end{itemize}

The selection of these 19 observational variables was carefully determined through extensive discussions with clinical anesthesiologists. These variables encompass crucial clinical information, vital signs, and the utilization of Analgesics, which directly influence the dosage administered by the anesthesiologist. Additionally, we have incorporated the concepts introduced in~\cite{schamberg2022continuous}, such as MAP target, MAP change, and MAP target error. These variables facilitate a faster perception of the target level, the subject's vital sign changes, and the true error for the agent. In the paper, the MAP target is defined as the average of the MAP of a patient over one operation.
$$
\text{MAP}^*_t=\frac{1}{T}\sum^T_{t=1}\text{MAP}_t,
$$
where $T$ is the duration of an operation and $\text{MAP}_t$ is the MAP of a patient at time step t. The MAP change is defined as the difference between the MAP at the current moment and the previous moment.
$$
\text{MAP}(change)_{t}=\text{MAP}_t-\text{MAP}_{t-1}.
$$
The MAP target error is defined as the distance between the current MAP and the target MAP.
$$
\text{MAP}(error)_{t}=\text{MAP}_t-\text{MAP}^*_t.
$$

\subsubsection{Action Space}
During automated anaesthetic infusion, more than one drug may be involved. In this study, the agent was only required to control propofol infusion rate. We also included the use of other drugs in the observed state to consider the synergistic effect of propofol with other drugs (\emph{e.g.}, analgesics). We calculated the maximum propofol dose used in the dataset $P_{\max}$ and performed a maximum normalization. The action space is thus $a\in \mathcal{A}$, continuous from 0 to 1. The final dose recommended by the agent is $a\cdot P_{\max}$, and this allows for safer infusion control.

\subsubsection{Reward Function}
During surgery, the primary goal of the agent is to ensure that the patient's vital signs remain within a specific range. In this study, we focused on MAP in particular since it is a readily accessible measurement and serves as an effective indicator of the patient's current vital status. We expected the agent to maintain the patient's MAP around the MAP target. After thorough consultations with clinical anesthesiologists, the reward function is designed as follows,
\begin{align*}
	R&_{error}(s_t,a_t)\\
	&=\begin{cases}
		+1 \quad &\text{if}~~|\text{MAP}_t-\text{MAP}^*_t |\le15\%\text{MAP}^*_t;\\
		+0.5 \quad &\text{if}~~15\%\text{MAP}^*_t\le|\text{MAP}_t-\text{MAP}^*_t |\le30\%\text{MAP}^*_t;\\
		-1 \quad &\text{else.}
	\end{cases}
\end{align*}
We adopted a segmented reward function design instead of a continuous one to enable the learning of more robust policies. In the surgical environment, there are numerous sources of noise and unpredictable events, such as intubation and sensor detachment. Utilizing a continuous reward function, such as mean squared error, could potentially cause the agent to overfit to these abnormal fluctuations, thereby increasing risks.


In addition, we hoped that the agent would use the lowest possible dosage to maintain the patient's vital signs within the target interval. To this end, we added a dose penalty as follows.
$$
R_{dosage}(s_t,a_t)=-\frac{|\text{MAP}_t-\text{MAP}^*_t|}{\text{MAP}^*_t}a_t,
$$
where $\frac{|\text{MAP}_t-\text{MAP}^*_t|}{\text{MAP}^*_t}$ is the adaptive correction factor. We do not want the agent to be overly conservative in the use of propofol, especially when the patient's MAP deviates from the target mean arterial pressure. Instead, the agent is expected to use a lower propofol dose to maintain the patient's vital signs within the target interval only when the patient's MAP is around the target. In summary, the overall reward function is calculated as follows.
\begin{equation}
	R_{total}(s_t,a_t)=R_{error}(s_t,a_t)+R_{dosage}(s_t,a_t).\label{reward}
\end{equation}

\subsection{Extending CQL As An Actor-critic Variant}
With the CQL constraint, we are able to learn a more reasonable Q-value estimation function by the Q-Learning method. However, Q-Learning may be too deterministic, especially during offline training, due to the presence of $\varepsilon$-greedy policy, which may limit the exploration of optimal actions by the agent. In this case, a stochastic policy may be a better choice.

To learn an explicitly parameterized stochastic policy, following~\cite{kumar2020conservative}, we first instantiate the CQL as an actor-critic algorithm.  In this framework, the CQL constraint term is first added to the training of the Q function and subsequently jointly trained with the actor as a critic. As in common implementations of actor-critic, an entropy term is added to the objective of the policy optimization in order to encourage policy exploration. Hereby, the gradient of the objective function of the actor can be formalized as
\begin{equation}
\begin{aligned} 
J(\pi_{\theta})=\mathbb{E}_{s\sim{D},a\sim{\pi_\theta}(a|s)}[\log\pi_{\theta}(a|s)Q(s,a)+\beta H(\pi_{\theta}(s))],
\end{aligned}\label{actor}
\end{equation}
where $H$ is the entropy of the policy and $\beta$ is the temperature factor used to relax the relative importance between the entropy term and reward.

\subsection{Policy Constraint Q-Learning}

In the aforementioned actor-critic framework, CQL serves as a constraint on the Q-function for accurate estimation of Q-values. However, during the iterative process of policy evaluation and improvement, errors in Q-value estimation can persist, potentially leading to an erroneous Q-function that drives the policy towards out-of-distribution (OOD) actions. To safeguard agent's actions, we add a constraint term to the CQL from a policy perspective. Our goal is to learn a function from a collected offline dataset to evaluate the plausibility of state-action pairs.


\begin{algorithm}[!ht]
	\caption{Policy Constraint Q-Learning (PCQL)}  
	\label{alg:PCQL}  
	\begin{algorithmic}[1]
		\STATE {\textbf{Input:} training set $\mathcal D$, number of training epochs $E$}  
		\STATE {Construct offline replay buffer $\mathcal D'$ from $\mathcal D$ using \eqref{reward}}
		\STATE {Initialize policy constraint network $\Phi_{\gamma}$, critic network $Q_{\theta}$ and actor network $\Pi_{\phi}$ with random parameters}
		\STATE {Initialize target network $Q_{\theta^{\prime}}$ with weights $\theta^{\prime}=\theta$}
		\FOR {epoch=1 \TO E} 
		\REPEAT 
		\STATE {Sample a mini-batch from replay buffer $\mathcal D'$} 
		\STATE {Train policy constraint $\Phi_{\gamma}$ by minimizing Eq. \eqref{cycle} and Eq. \eqref{entropy}}
		\STATE {Train critic $Q_{\theta}$ by minimizing Eq.~\eqref{cql}}
		\STATE {Train actor $\Pi_{\phi}$ by maximizing Eq.~\eqref{total}}
		\STATE {Every $N$ steps update target network with $\theta^{\prime}=\theta$}
		\UNTIL{all mini-batch are sampled}
		\ENDFOR
		\STATE {\textbf{Output:} optimal PCQL policy $\Pi_{\phi}^*$}
	\end{algorithmic}  
\end{algorithm} 

\subsubsection{Policy Constraint}
The scoring function consists of two components, the environment transition model $h$ and the behavior prediction model $g$. The environmental transition model aims to predict the new state to which the current state $s$ will be transferred after the selection of action $a$, i.e., $s'=h(s,a)$. And the behavior prediction model predicts the possible actions that may be taken  given two consecutive states, i.e., $a=g(s,s')$. For the action $\hat a=f(s)$ predicted by the agent at the current state $s$ (\emph{i.e.}, the dose), we first imagine its subsequent state using the environmental transition model, and then reason about the possible action $\dot a$ using the behavioral prediction model, i.e., $\dot a=g(s,h(s,\hat a))$. We constrain the action $\hat a$ predicted by the agent not to be too far from the action $\dot a$ inferred by the behavioral prediction model as likely to occur. A simple approach is to
$$
\Phi_{\text{simple}}(s,\hat a) =\|g(s,h(s,\hat a))- \hat a\|_2,
$$
where $\|\cdot\|_2$ denotes the Euclidean distance.

However, such a deterministic constraint leads to the fact that the actions predicted by the agent are supervised by the output of the behavior prediction model. This is perhaps too stringent, thus weakening the exploration of other, more optimal actions. For this reason, we propose an alignment constraint in the latent space. The network architectures of the environment transition model $h$ and the behavior prediction model $g$ are similar, and both consist of an encoder, a predictor, and a projector. The predictor aims to predict the target information (\emph{e.g.} $s'$ for $h(s,a)$), while the projector is responsible for mapping the output of the encoder to the latent space, a process we denote as $\mathbf {Prj}(\cdot)$. We use the cross entropy function for alignment constraints, i.e.
$$
\begin{aligned}
\Phi(s,\hat a) =&\\
\text{softmax}&\left(\frac{\mathbf {Prj}(h(s,\hat a))}{\tau}\right)\cdot\log\left(\text{softmax}\left( \frac{\mathbf {Prj}(h(s,\dot a))}{\tau}\right)\right),
\end{aligned}
$$
where $\tau$ is the temperature factor for scaling the learning difficulty. Combining Eq.~\eqref{actor}, the final objective function of actor is
\begin{equation}
	L_{total}=J(\pi_{\theta})+\Phi(s,\hat a).\label{total}
\end{equation}


\subsubsection{Training Strategy}
We next discuss the training of the environment transfer model $h$ and the behavior prediction model $g$. In fact, both models can be learned from an offline dataset following a supervised learning paradigm. We sample $(s,a,s')\sim\mathcal D$ and train these two models using an autocoder-like approach, i.e., 
\begin{equation}
	L_{cycle}=(g(s,h(s,a)) - a) + (h(s,g(s,s')) - s'). \label{cycle}
\end{equation}
Also, we add the entropy consistency as follow,
\begin{equation}
	L_{entropy}=\mathcal H(h(s,\dot a), h(s,a))+\mathcal H(g(s,\dot s'), g(s,s')),\label{entropy}
\end{equation}
where $\mathcal H(a,b)=-\text{softmax}(a/\tau)\cdot\log(\text{softmax}(b/\tau))$ and $ \tau$ is the temperature factor. We jointly train the scoring function $\Phi(s,a)$ and the agent of the CQL algorithm, while adding consistency constraints to the action loss of CQL. The pseudo-code of the overall training algorithm of the proposed PCQL is shown in Algorithm \ref{alg:PCQL}.

\section{Experiments and Results}
\label{sec:exp}

\subsection{Data Collection and Training Details}
\begin{figure}[!ht] 	 	
	\centerline{\includegraphics[width=0.95\columnwidth]{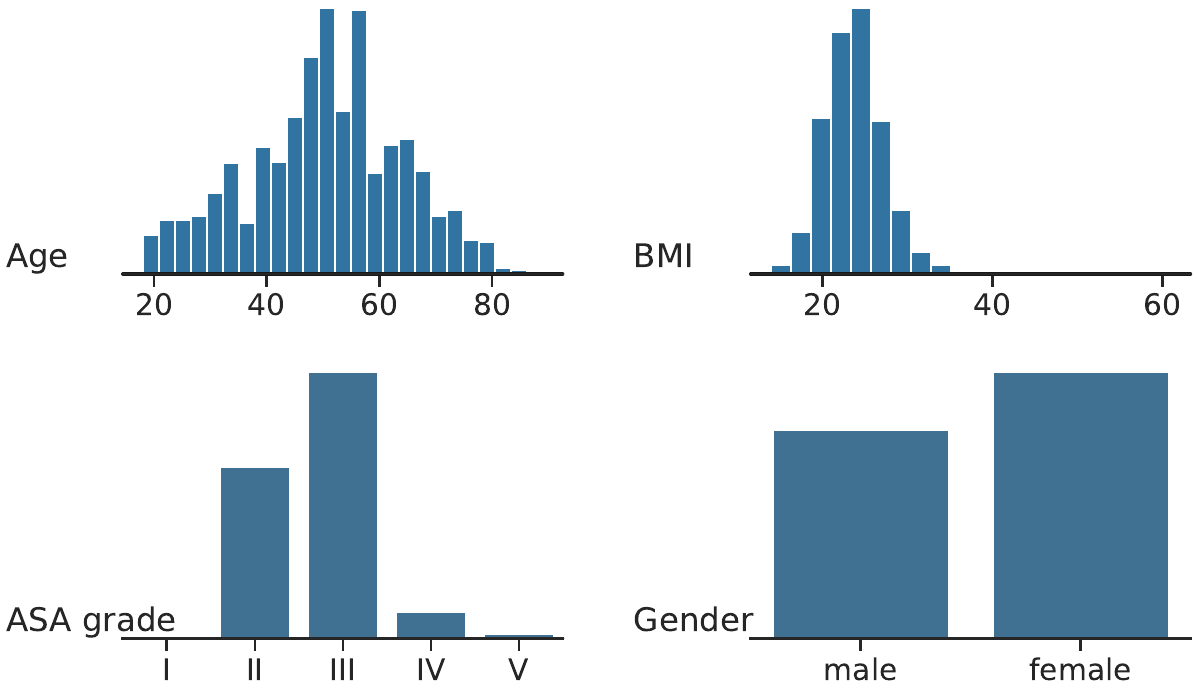}} 	 	
	\caption{Clinical data distributions of collected dataset.} 	 	
	\label{fig:pop}  
\end{figure}

We collected 6,303 brain-neurological surgeries involving general anesthesia from West China Hospital. During each surgery, the patient's vital sign signals as well as the infused drugs information was recorded in real time at one-minute intervals. We filtered the surgeries that met any of the following conditions: i) missing corresponding dosing information; ii) surgical time shorter than half an hour; iii) severe missing vital sign records during surgery; and iv) samples using inhaled anesthesia, such as sevoflurane or desflurane instead of propofol. We used the k-NN method to fill in the remaining missing values. Finally, a total of 1,293 surgeries with 284,281 anesthesia records were obtained. Relevant clinical data are summarized in Fig.~\ref{fig:pop}. All experiments are conducted based on d3rlpy~\cite{d3rlpy}. We employed five-fold cross-validation in order to ensure a reliable evaluation of models. We use Adam~\cite{kingma2014adam} as the optimizer and train for 200 epochs with batchsize 256. The temperature factor for entropy consistency is set to 0.08. The learning rates of actor, critic, environment transition model, and action prediction model are set to $1\times10^{-4}$, $3\times10^{-4}$, $1\times10^{-4}$ and $3\times10^{-4}$ respectively. These hyperparameter settings are obtained from the grid search to ensure the performance and generalization of the agent. Specifically, the learning rate is searched in $\{1\times10^{-3}, 3\times10^{-3}, 1\times10^{-4}, 3\times10^{-4}\}$ while the temperature factor is searched in $\{0.03, 0.05, 0.08\}$.

\subsection{Evaluation Protocol}

\subsubsection{Baselines}
We mainly set the following baselines, including anesthesiologists' clinical policy (ANE), soft actor-critic (SAC,~\cite{haarnoja2018soft}), batch constrained deep Q-learning (BCQ,~\cite{le2019batch}), conservative Q-learning (CQL,~\cite{kumar2020conservative}), twin delayed deep deterministic policy gradient with behavior cloning (TD3+BC,~\cite{fujimoto2021minimalist}) and mildly conservative Q-Learning (MCQ,~\cite{lyu2022mildly}). ANE represents the actual clinical practice of anesthesiologists in the collected dataset and serves as our primary reference of recommended dose. To demonstrate the advantages of offline reinforcement learning in the absence of an available environment, we include a popular off-policy RL algorithm, SAC, in our comparisons. Furthermore, to highlight the strengths of PCQL, we compare it with several state-of-the-art offline RL techniques, including BCQ, CQL, TD3+BC, and MCQ. BCQ aims to minimize the distance between selected actions and actions in the dataset by employing a conditional VAE on the state. TD3+BC achieves policy constraints by simply adding a behavioral cloning term to the TD3 algorithm\cite{fujimoto2018addressing}. MCQ is an improved algorithm for CQL by assigning pseudo target values to actively train OOD actions.


\subsubsection{Off Policy Evaluation}
In our experimental settings, no interactive environment is available, so trained policies cannot directly compute cumulative rewards by interacting with the environment. We used the off-policy evaluation (OPE) method commonly used in ORL for performance evaluation~\cite{voloshin2019empirical}, which was done on a divided test set. In recent experiments in healthcare settings, researchers have found that the Fitted Q Evaluation (FQE;~\cite{le2019batch}) approach consistently yields accurate policy performance evaluation results~\cite{tang2021model, kondruppersonalization, zhu2023offline}, and we follow this practice. FQE uses a fixed trained policy, and a re-set Q function, and then re-trains the Q function. The retrained Q function estimates how much return the trained policy can obtain from initial states. The higher the estimated returns, the better the expected performance of the policy. To comprehensively evaluate the performance of policies, we also use Soft Off-Policy Classification (Soft OPC;~\cite{irpan2019off}) as another supplementary metric. Soft OPC is a policy-based evaluation metric that measures algorithm performance by comparing action values between successful and unsuccessful trajectories. If the action values in successful trajectories are significantly higher than those in unsuccessful trajectories, it indicates better policy performance of the algorithm.

\begin{figure*}[!ht]
	\centerline{\includegraphics[width=1\linewidth]{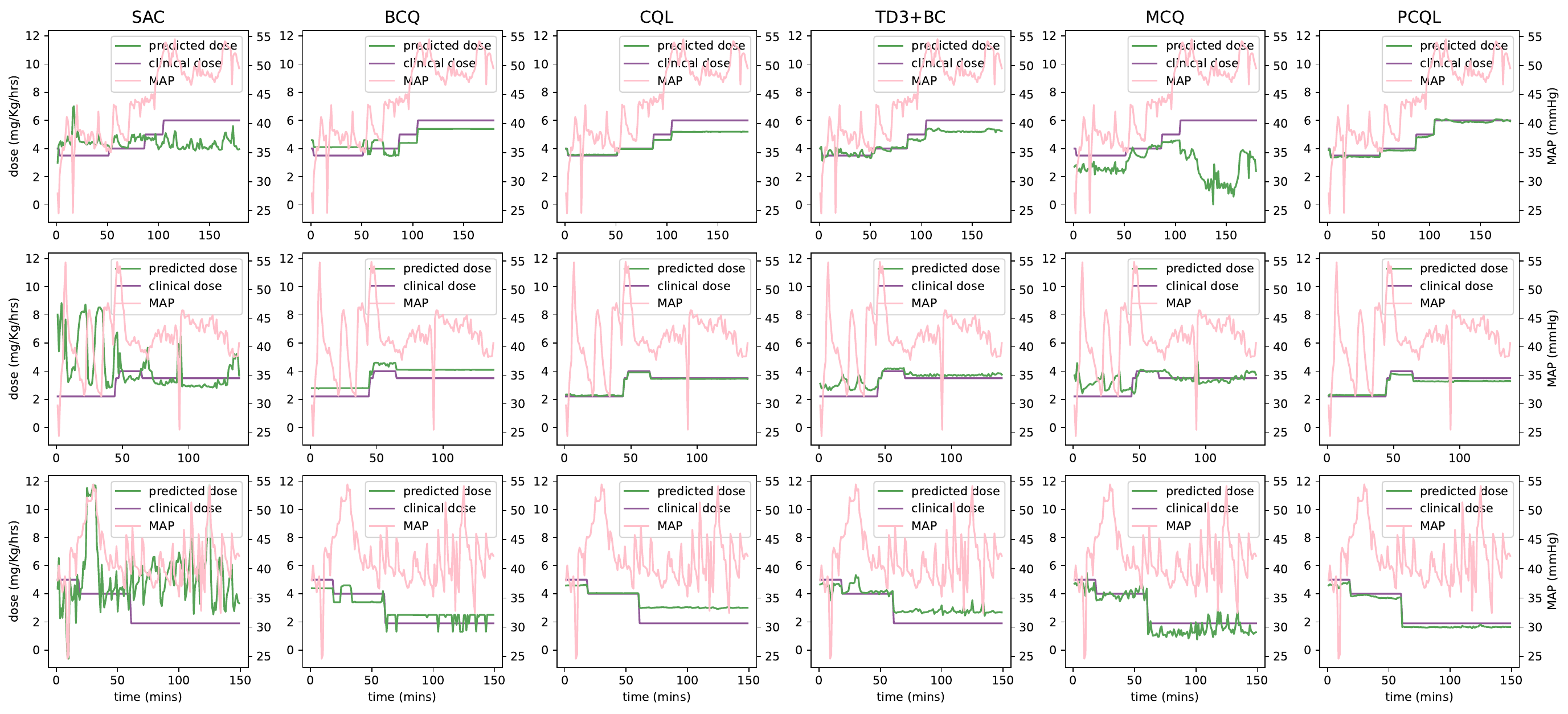}}
	\caption{Comparison of recommended doses in general anesthesia cases among different algorithms and anesthesiologists (each row represents a case).}
	\label{fig:cmp_pred_gt_map}
\end{figure*}

	\begin{table*}[!h]
		\caption{Comparison of the predicted returns of different methods}
		\label{tab:init_state}
		\centering
		\begin{tabular}{cccccccc}
			\toprule
			~ & ANE & SAC~\cite{haarnoja2018soft} & BCQ~\cite{le2019batch} & CQL~\cite{kumar2020conservative} & TD3+BC~\cite{fujimoto2021minimalist} & MCQ~\cite{lyu2022mildly} & PCQL \\
			\midrule
			\textbf{Initial state $\uparrow$} & 59.071$\pm$0.663 & 43.325$\pm$12.004 & 64.919$\pm$2.099 & 66.598$\pm$1.111 & 66.693$\pm$1.506 & 69.043$\pm$1.714  & \textbf{73.069}$\pm$\textbf{1.302} \\
			\textbf{Soft OPC $\uparrow$} & 0.072$\pm$0.002 & -1.274$\pm$1.015 & -0.184$\pm$2.974 & 0.633$\pm$0.388 & -0.394$\pm$1.814 & 0.434$\pm$0.879  & \textbf{1.061}$\pm$\textbf{0.416} \\
			\bottomrule
		\end{tabular}
	\end{table*}

\begin{table*}[!h]
	\caption{Performance comparison in retrospective evaluation}
	\label{tab:error}
	\centering
	\begin{tabular}{ccccccc}
		\toprule
		~ & SAC~\cite{haarnoja2018soft} & BCQ~\cite{le2019batch} & CQL~\cite{kumar2020conservative} & TD3+BC~\cite{fujimoto2021minimalist} & MCQ~\cite{lyu2022mildly} & PCQL \\
		\midrule
		\textbf{MAPE (\%)} $\downarrow$ & 78.557$\pm$30.034 & 20.510$\pm$11.236 & 8.884$\pm$2.435 & 9.762$\pm$1.822  & 12.214$\pm$5.486 & \textbf{7.860}$\pm$\textbf{1.475} \\
		\textbf{RMSE} $\downarrow$ & 3.522$\pm$1.270 & 0.800$\pm$0.497 & 0.323$\pm$0.096 & 0.390$\pm$0.053  & 0.437$\pm$0.087  & \textbf{0.279}$\pm$\textbf{0.068} \\
		\bottomrule
	\end{tabular}
\end{table*}

\subsubsection{Retrospective Evaluation}
Since the data we collect is retrospective from the operating room, we can also evaluate the model in a "consult" mode. In this setting, the intelligence can see the historical state and action information, and then give the recommended dose for the current state. At the same time, we take the actual clinical dose given by the anesthesiologist as the optimal dose and calculate the difference between the recommended dose and the actual dose. Although we could not know how the human body responds to the recommended dose by the agent, we argue  that this evaluation can still in part reflect the performance of the agent when faced with the real world.  To measure the discrepancy between the recommended doses and the actual ones, we used two common regression metrics for evaluation: mean absolute percentage error (MAPE) and root mean square error (RMSE). In this case, MAPE is defined as
$$
\frac{1}{NT}\sum_{i=1}^{N}\sum_{t=1}^{T}\frac{|y_{i,t}-y_{i,t}^*|}{\max(\varepsilon,y_{i,t}^*)}100\%
$$
where $N$ is the number of episodes and $T$ is the length of an episode. $y_{i,t}$, $y_{i,t}^*$ denote the recommended dose of the agent and the actual dose for the $i$ episode at the $t$ moment, respectively. The $\varepsilon$ stands for a very small number to avoid division by zero error. RMSE is defined as
$$
\frac{1}{NT}\sum_{i=1}^{N}\sum_{t=1}^{T}\sqrt{(y_{i,t}-y_{i,t}^*)^2}
$$

\subsection{Quantitative Analysis of Performance}



In this section, we compare the PCQL with baselines, including ANE, BCQ, CQL, TD3+BC and MCQ. First, we evaluate the policies on the initial state and Soft OPC metrics, and the quantitative results are presented in Table~\ref{tab:init_state}. We can see that PCQL is expected to yield higher returns than the other policies. The estimated return of PCQL is 1.24 times higher than that of the anesthesiologist's strategy and also superior to MCQ (73.069 compared to 69.043). Moreover, PCQL achieved the highest Soft OPC score, suggesting that the actions recommended by PCQL are more likely to outperform the actions taken by the behavior policy. It is worth noting that all the RL-based policies perform well except SAC. One possible reason for its poor performance is that the off-policy RL algorithm cannot correct the value estimation errors of the OOD actions in time due to the unavailability of the environment. Consequently, the extrapolation errors continue to propagate during training, ultimately leading to the training failure. In contrast, the ORL algorithms mitigate this issue by constraining the policy distribution either Q-value estimation. PCQL combines their ideas and achieves more advantageous results.

\begin{figure*}[htbp]
	\centering
	\begin{minipage}[t]{0.48\textwidth}
		\centering
		\centerline{\includegraphics[width=0.95\columnwidth]{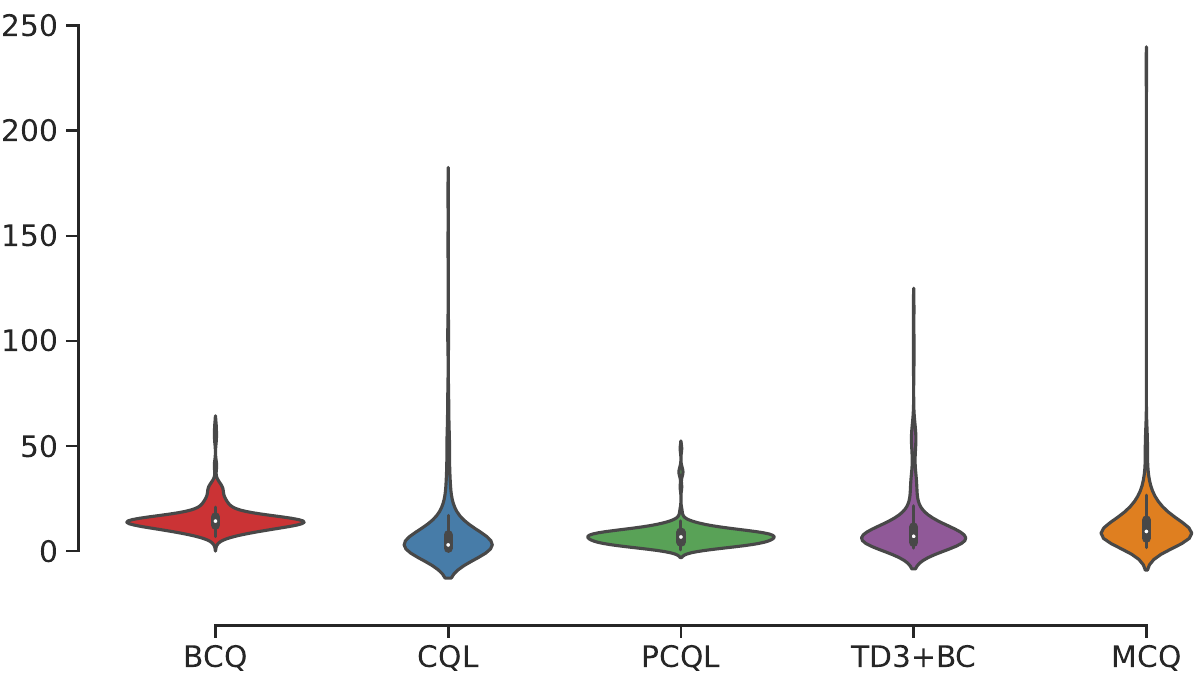}} 	
		\caption{Comparison of different offline RL algorithms in terms of absolute percentage error (APE) metric.} 	
		\label{fig:mape}
	\end{minipage}
	\begin{minipage}[t]{0.48\textwidth}
		\centering
		\centerline{\includegraphics[width=0.95\columnwidth]{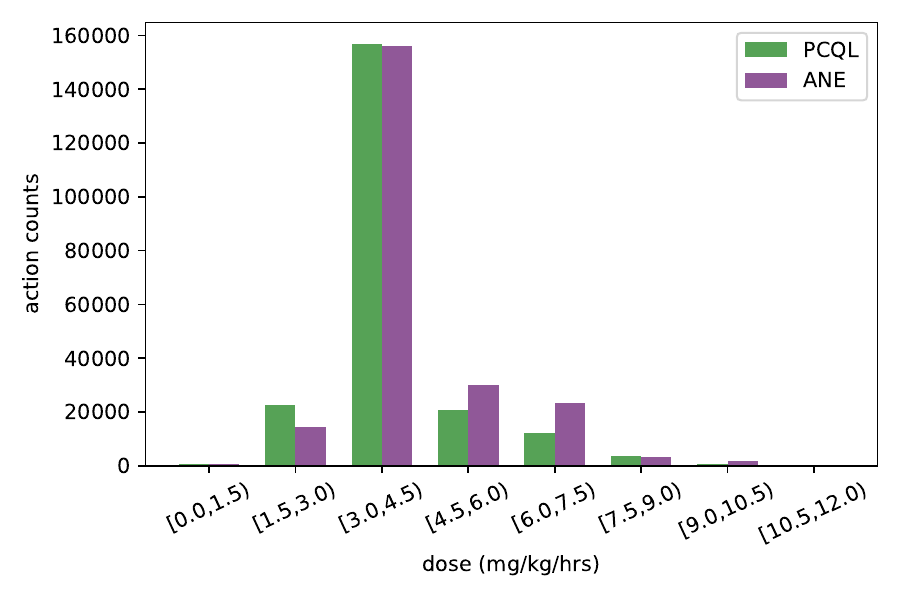}} 	
		\caption{Comparison of the distribution of doses used by PCQL and anesthesiologists.} 	
		\label{fig:dose_dist}
	\end{minipage}
\end{figure*}

\begin{table*}[!h]
	\centering
	\begin{minipage}{0.45\textwidth}
		\centering
		\small
		\makeatletter\def\@captype{table}\makeatother
		\caption{Comparison of mean doses used}
		\label{tab:mean_dose}
		\begin{tabular}{ccc}
			\toprule
			~ & ANE & PCQL \\
			\midrule
			\textbf{Mean dose (mg/Kg/hrs)} & 3.921 & 3.719 \\
			\bottomrule
		\end{tabular}
	\end{minipage}\quad
	\begin{minipage}{0.45\textwidth}
		\centering
		\small
		\makeatletter\def\@captype{table}\makeatother
		\caption{Comparison of the correlation}
		\label{tab:corr}
		\begin{tabular}{ccc}
			\toprule
			~ & ANE & PCQL \\
			\midrule
			\textbf{Correlation} & 0.081 & 0.245 \\
			\bottomrule
		\end{tabular}
	\end{minipage}
	\vspace{-15pt}
\end{table*}

Subsequently, we evaluated the trained models in "consultation" mode, focusing on the discrepancy between the recommended dose from different policies and the actual clinical dose. To compare the performance of the policies, we used two common metrics, namely MAPE and RMSE, and present the results in Table~\ref{tab:error}. As shown in the result, PCQL consistently demonstrates superior and robust performance. While BCQ achieved a less favorable result (20.510\%) on the MAPE metric, PCQL, TD3+BC, and CQL were all within 10\% of each other. Although MCQ showed promising performance on the initial state metric, it only achieved a 12.214\% result on the MAPE metric. Notably, SAC, the traditional RL approach, performed poorly on both metrics, highlighting the importance of incorporating certain constraints when the environment is not available.

We further analyzed the MAPE performance of different ORL methods in Fig.~\ref{fig:mape}. Our results show that the absolute percentage error (APE) of PCQL had a lower overall distribution, and thanks to the consistency constraint, the APE of PCQL exhibited reduced variance and fewer outliers compared to the baselines. Finally, we visualized the performance of different policies in several randomly selected cases from the test set. Overall, doses recommended by PCQL aligned well with the clinical doses of the anesthesiologists. In contrast, SAC showed poor fluctuations. Other ORL algorithms also aligned with the clinical doses but exhibited different local oscillations (e.g., BCQ in the third row of Fig.~\ref{fig:cmp_pred_gt_map}) and deviations (e.g., MCQ in the first row of Fig.~\ref{fig:cmp_pred_gt_map}).

\subsection{Model Recommended Dosage Analysis}
Given that PCQL significantly outperforms baselines in both previous performance evaluations, we next focus exclusively on analyzing PCQL's recommended dose usage and comparing its advantages and disadvantages with the anesthesiologist's policy. 


The statistical results of mean dose values on the test set are shown in Table \ref{tab:mean_dose}. We found that the mean dose value of 3.719 mg/Kg/hrs recommended by PCQL is slightly lower than the actual clinical dose of 3.921 given by the anesthesiologist by 0.202 mg/Kg/hrs. The difference of 0.202 mg/Kg/hrs reduction is clinically acceptable, because even two experienced anesthesiologists may give a difference of 1 mg/Kg/hrs in the anesthetic dose for the same case of anesthetic surgery. At the same time, it is clinically meaningful if the PCQL policy is able to maintain the patient's depth of anesthesia within the target range using a lower drug dose. First, it would mean that the problem of over-anesthesia could be alleviated, reducing the incidence of post-anesthesia syndrome and thus improving the quality of patients' surgery. Second, lower dose use often means less consumption of medical resources, which helps reduce the cost of patient care. As a supplement, we also present the distributional difference of doses recommended by PCQL and human anesthesiologists in the test set (as depicted in Fig.~\ref{fig:dose_dist})


Intriguingly, we noted that the dose recommended by PCQL fluctuated around the actual dose given by the anesthesiologist and that this fluctuation was positively correlated with the MAP of the patient (for instance, MCQ in the first row of Fig.~\ref{fig:cmp_pred_gt_map}). To verify this, we calculated Pearson correlation coefficients between recommended doses and MAPs for PCQL and anesthesiologist policies, respectively. The Pearson correlation coefficient was calculated by
$$
\rho_{X,Y}=\frac{\mathbb{E}{((X-\mu_X)(Y-\mu_Y))}}{\sigma_X\sigma_Y},
$$
where $\mu_X$ and $\sigma_X$ denote the mean and variance of the variable $X$, respectively; similarly for $Y$.



As indicated in Table~\ref{tab:corr}, the dose recommended by PCQL showed a stronger correlation with MAP compared to anesthesiologists' (0.245 compared to 0.081). This is probably because in clinical practice, anesthesiologists are typically responsible for multiple procedures and cannot attend to the same patient continuously. Furthermore, the focus of human anesthesiologists cannot be sustained for extended periods. In contrast, as illustrated in Fig.~\ref{fig:cmp_pred_gt_map}, the dose recommended by PCQL are more sensitive and responsive to changes in the patient's vital signs. This suggests the potential for computer-assisted precision anesthesia and offers the prospect of delivering higher quality, more personalized anesthesia infusion management.

\subsection{Confidence Interval Analysis of the Model}

\begin{figure}[t] 	
	\centerline{\includegraphics[width=0.95\columnwidth]{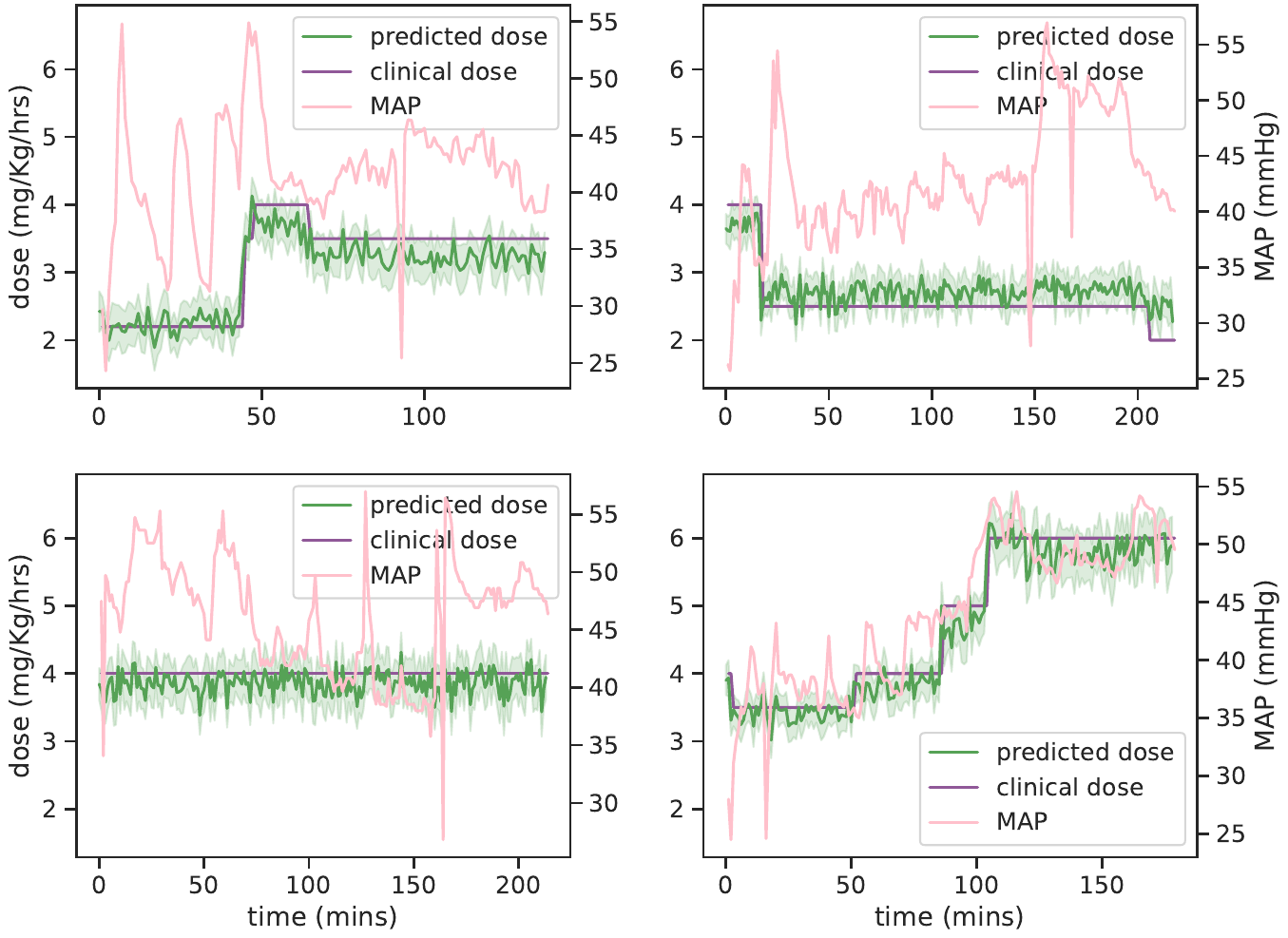}} 	
	\caption{Confidence interval estimation of PCQL in general anesthesia cases.} 	
	\label{fig:uncertainty} 
\end{figure}

It is a fact that even experienced anesthesiologists cannot determine the optimal dose\textendash they usually pick a preferred dose within a confidence interval that they believe is reasonable. To estimate the confidence interval for the PCQL, we simply modified the output mechanism of the PCQL so that it probabilistically samples an action when making inferences, rather than the best one. We choose the common Gaussian policy of $\pi_\theta(a|s)=\frac{1}{\sqrt{2\pi}\sigma}\exp\left({-\frac{a-f_\theta(s)}{2\sigma^2}}\right)$. At state $s$, the actions taken when sampled by this policy obey a normal distribution with a mean of $f_\theta(s)$ and a variance of $\sigma^2$. Fig. \ref{fig:uncertainty} gives the results of sampling 100 times, and it can be seen that the confidence interval estimated by PCQL is on the safe side without any serious deviation from the actual clinical dose. The confidence interval recommended by PCQL covers the majority of the anesthesiologist's clinical policy, and this is encouraging. Because it means that under this situation, the anesthesiologist's policies can be derived from sampling within the confidence interval of the PCQL. Although we could not be informed in retrospective data about the confidence intervals considered by anesthesiologists, we argue that this enhances the credibility and reliability of PCQL.

\subsection{Interpretability of the Model}

The deep neural networks used to implement our policy are notorious for their black-box characteristics. However, if an anesthesiologist does not know how the machine model makes decisions, then he will not be able to trust the use of the model. To increase the transparency of the model, we analyzed the prediction results of the model using SHapley Additive exPlanations (SHAP;~\cite{lundberg2017unified}). SHAP is a game-theoretic interpretation approach that predicts the output of a model by fitting a linear interpretation model, and calculates SHAP values for observed features, which indicate the importance of the features in influencing the model output. SHAP values have been widely applied to analyze the relative importance of features in various machine learning models for medical applications~\cite{schamberg2022continuous, lundberg2018explainable, tjoa2020survey}. We used the publicly available SHAP library\footnote{https://github.com/slundberg/shap} to calculate the SHAP values for each feature in the observation space, and obtained the Absolute Mean SHAP Score for each feature by taking the average of the absolute values. 

\begin{figure}[t] 	
	\centerline{\includegraphics[width=0.95\columnwidth]{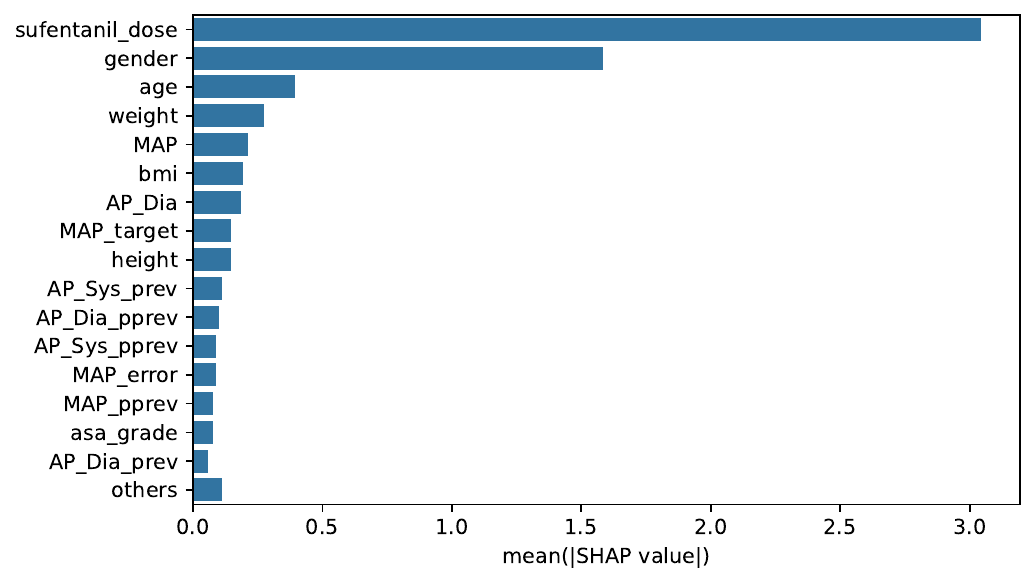}} 	
	\caption{Results of absolute mean SHAP values for different input components of PCQL.} 	
	\label{fig:barh} 
\end{figure}

As shown in Fig. \ref{fig:barh}, we found that the output of PCQL (\emph{i.e.}, the infusion rate of propofol) was most affected by sufentanil dose. We attribute this to the fact that sufentanil is an immediate injection analgesic that rapidly depresses the patient's respiration and, in addition, acts synergistically with propofol, thus greatly influencing the output of the model. Clinical characteristics of patients such as gender, age, and weight also significantly influenced model output, followed by real-time vital sign characteristics such as MAP, $\text{AP}_\text{Dia}$, and MAP target. We believe this is similar to the clinical decision making of the anesthesiologist\textendash the anesthesiologist usually determines the primary base dose by pre-assessing the patient's physical condition, and adjust the base dose according to changes in the patient's vital signs during surgery.

\section{Conclusion}
In this study, we proposed PCQL, an ORL algorithm for automated anesthesia infusion. To ensure the safety of the dosing policy, we have imposed constraints on both the policy distribution and value estimation. Through extensive experiments on a large collected anesthesia dataset, we validated the effectiveness of PCQL. The experimental results demonstrated that our method outperformed baseline methods. The recommended dose generated by PCQL was consistent with the anesthesiologist's strategy, while utilizing a lower total dose and being more responsive to the patient's vital signs. This highlights the potential for future personalized and precise anesthesia. Additionally, we employed SHAP to perform interpretability analysis on PCQL's results, enhancing the transparency of our approach.

However, it is important to acknowledge that further validation through real-world clinical experiments is necessary to fully assess the effectiveness of PCQL. Our study represents a promising step towards achieving true automation in anesthesia, and we recognize the need for continued research and refinement in this area. We hope that our findings will inspire further advancements in the field to ultimately contribute to improved patient care.


\section{Limitations and Future Work}
Our study confirms the practicality of using ORL for automated anesthesia, although there are several potential concerns. Firstly, our data is limited to a single center, which does not allow us to assess PCQL's performance in a multicenter context. In future work, we intend to gather clinical anesthesia data from multiple centers and evaluate the model's effectiveness accordingly. For this purpose, an appropriate combination of federated learning~\cite{ali2021integration, ali2022federated, zhou2022federated} and ORL might be a compelling direction. Secondly, for off-policy evaluation, we employed FQE and Soft OPC as our evaluation protocols. While there is ample literature demonstrating the reliability and credibility of these metrics, and PCQL has shown favorable results based on these metrics, there is still a lack of comprehensive comparative experiments that establish the alignment of these evaluation metrics with real-world performance. It is crucial to ensure that the results of OPE accurately reflect the clinical outcomes observed in real-world scenarios. Once this alignment is established, it may be possible to deploy ORL-trained agents in real clinical settings under controlled conditions. Further research is warranted to explore the design of OPE in the specific context of healthcare.

\bibliographystyle{IEEEtran}
\bibliography{generic-color}{}

\end{document}